%% file: main.tex
\documentclass[11pt,a4paper]{article}
\usepackage[hyperref]{acl2020}
\usepackage{times}
\usepackage{makecell}
\usepackage{latexsym}

\usepackage{soul}
\usepackage{enumitem}

\usepackage{amsmath}
\usepackage{amsfonts}
\usepackage{bm}
\usepackage{graphicx}
\usepackage{subfigure}
\usepackage{multirow, booktabs}
\usepackage{siunitx}
\usepackage{array}
\usepackage{color}
\usepackage{booktabs}
\usepackage{colortbl}
\usepackage{mathrsfs}

\usepackage{amsmath}
\usepackage[linesnumbered,ruled]{algorithm2e}

\definecolor{CaseColor1}{RGB}{255,140,0}
\definecolor{CaseColor2}{RGB}{205,38,38}
\definecolor{CaseColor3}{RGB}{46,139,87}

\usepackage{longtable}

\newtheorem{myDef}{Definition} 
\newtheorem{myFor}{Formulation}

\usepackage{microtype}

\aclfinalcopy % Uncomment this line for the final submission
\def\aclpaperid{3009} %  Enter the acl Paper ID here

\renewcommand{\thefootnote}{\fnsymbol{footnote}}

\title{TAG : Type Auxiliary Guiding for Code Comment Generation}

\author{
Ruichu Cai$^1$, 
Zhihao Liang$^1$, 
Boyan Xu$^1$\thanks{$^*$ Corresponding author} ,
Zijian Li$^1$,
Yuexing Hao$^2$,
Yao Chen$^3$
\\
$^1$ School of Computer Science, Guangdong University of Technology, China \\
$^2$ Rutgers University New Brunswick, USA  \\
$^3$ Advanced Digital Sciences Center, Singapore\\
cairuichu@gmail.com,
zhihaolzh95@gmail.com,
hpakyim@gmail.com,\\
leizigin@gmail.com,
yh599@scarletmail.rutgers.edu,
yao.chen@adsc-create.edu.sg
}

\begin{document}

\maketitle
\renewcommand{\thefootnote}{\arabic{footnote}}
\begin{abstract}
% Code comment generation, aiming at extracting the source code into human-understandable description, is an important but challenging natural language generation task. 
Existing leading code comment generation approaches with the structure-to-sequence framework ignores the type information of the interpretation of the code, e.g., operator, string, etc.
However, introducing the type information into the existing framework is non-trivial due to the hierarchical dependence among the type information.
In order to address the issues above, we propose a Type Auxiliary Guiding encoder-decoder framework for the code comment generation task which considers the source code as an N-ary tree with type information associated with each node.
Specifically, our framework is featured with a \emph{Type-associated Encoder} and a \emph{Type-restricted Decoder} which enables adaptive summarization of the source code.
% \textcolor{blue}{\st{by adjusting the parameters of the Tree-LSTM according to the current node's type and reduces the search space of the decoding process through a type-restricted two-stage generation process, respectively.}} 
We further propose a hierarchical reinforcement learning method to resolve the training difficulties of our proposed framework.
Extensive evaluations demonstrate the state-of-the-art
% \textcolor{blue}{benefits} of type information in the code comment generation tasks and \textcolor{blue}{demonstrate the effectiveness of our framework.}
performance of our framework with both the auto-evaluated metrics and case studies.
\end{abstract}

\input{sections/01_intro_3.tex}

\input{sections/02_related.tex}
\input{sections/03_overview_2.tex}

\input{sections/04_model_2.tex}

\input{sections/05_typeDecoder.tex}
\input{sections/06_rl.tex}

\input{sections/07_exp.tex}

\input{sections/08_conclusion.tex}

\input{sections/acknowledgement}

\bibliography{acl2020}
\bibliographystyle{acl_natbib}

%\appendix

% \input{sections/appendix.tex}
%\label{sec:appendix}

\end{document}

\section{Appendix A}
\subsection{Information of Datasets}

\begin{table*}[htb]
	\centering
	\small
	\begin{tabular}{|p{40pt}<{\centering}|p{20pt}<{\centering}|p{25pt}<{\centering}|c|c|c|c|c|c|c|c|c|}
		\hline
		Dataset&\textbf{Type Num}&\textbf{Avail. Types Num}&
		\multicolumn{3}{c|}{\textbf{Max Tree Depth}}&\multicolumn{3}{c|}{\textbf{Avg Tree Node Num}}&
		\multicolumn{3}{c|}{\textbf{Max Child Num}}\\
		\hline
		\multirow{2}{40pt}{WikiSQL}&\multirow{2}{5pt}{6}&\multirow{2}{5pt}{2}
		&train&dev&test&train&dev&test&train&dev&test\\
		\cline{4-12}
		&&&5&5&5&11.09&11.14&11.17&4&4&4 \\
		\hline
		\multirow{2}{40pt}{ATIS}&\multirow{2}{5pt}{7}&\multirow{2}{5pt}{5}
		&train&dev&test&train&dev&test&train&dev&test\\
		\cline{4-12}
		&&&18&16&14&33.90&34.01&29.54&15&8&9 \\
		\hline
		\multirow{2}{40pt}{CoNaLa}&\multirow{2}{5pt}{22}&\multirow{2}{5pt}{6}
		&train&dev&test&train&dev&test&train&dev&test\\
		\cline{4-12}
		&&&28&24&26&28.20&27.46&29.48&10&6&9 \\
		\hline
	\end{tabular}
	\caption{Information of trees over the three data sets}
	\label{info-table}
\end{table*}

\begin{table*}[htb]
	\centering
	\begin{tabular}{|p{80pt}<{\centering}|p{130pt}<{\centering}|p{120pt}<{\centering}|}
	%\begin{tabular}{|c|c|c|}
		\hline
		Dataset&Types&Avail. Types in Dec.\\
		\hline
		WikiSQL&stmt, agg\_op, column\_name, cond\_expr, cmp\_op, string&column\_name, string\\
		\hline
		ATIS&expr, var, var\_type, ent, num, pred, cmp\_op&var, ent, num, var\_type, pred\\
		\hline
		CoNaLa&object, arg, string, excepthandler, cmpop, int, withitem, singleton, unaryop, stmt, 
		identifier, arguments, comprehension, boolop,
		bytes, keyword, mod, slice, alias, expr, operator, expr\_context \footnote{the python types from the ASDL grammar of python, which can be searched on the python official website https://docs.python.org/3.5/library/ast.html}&identifier, int, string, bytes, object, singleton\\
		\hline
	\end{tabular}
	\caption{Types and available types in Type-Restricted Decoder (Avail. Types in Dec.) in data sets}
	\label{type-table}
\end{table*}

Table \ref{info-table} shows the statistic of trees over the three data sets, where \textbf{Type Num} denotes the number of types in each data set, \textbf{Avail. Types Num} denotes the number of available types in Type-restricted Decoder, \textbf{Max Tree Depth} denotes the maximum depth of trees over each data set, \textbf{Avg Tree Node Num} denotes the average number of tree nodes each tree over the data sets, \textbf{Max Child Num} denotes the maximum number of child nodes in each node of data sets. The details of types and available types are presented in Table \ref{type-table}. 

As shown as Table \ref{type-table}, we can find that the type contains rich information of the code. For example, there are types in column ``Types" of  Table \ref{type-table} for an AST of a SQL query on WikiSQL. And type ``column\_name" and type ``string", which are in column ``Avail. Types in Dec.", are allowed to be considered in Type-restricted Decoder.

\newpage
\section{Appendix B}
\subsection{Examples}

\begin{table}[htb]
	\begin{tabular}{|p{100pt}<{\centering}|p{300pt}<{\centering}|}
		\hline
		\multirow{8}{100pt}{\textbf{SQL}:SELECT MAX(Capacity) FROM table WHERE Stadium = "Otkrytie Arena"}
		&WikiSQL\\
		\cline{2-2}
		&\textbf{GOLD}:What is the maximum capacity of the Otkrytie Arena stadium ?\\
		\cline{2-2}
		&\textbf{Code-NN}:What is the highest attendance for ?\\
		\cline{2-2}
		&\textbf{P-G}:Who is the \% that ' s position at 51 ?\\
		\cline{2-2}
		&\textbf{Tree2Seq}:What is the highest capacity at <unk> at arena ?\\
		\cline{2-2}
		&\textbf{Graph2Seq}:What is the highest capacity for arena arena?\\
		\cline{2-2}
		&\textbf{T2S+CP}:What is the highest capacity for the stadium ?\\
		\cline{2-2}
		&\textbf{TAG}:What is the highest capacity for the stadium of Otkrytie Arena ?\\
		\hline
		\multirow{8}{100pt}{\textbf{lambda}:lambda \$0 e ( and ( flight \$0 ) ( or ( class\_type \$0 first:cl ) ( class\_type \$0 coach:cl ) ) ( from \$0 ap0 ) ( to \$0 ci0 ) )}
		&ATIS\\
		\cline{2-2}
		&\textbf{GOLD}:show me first class and coach flight between ap0 and ci0\\
		\cline{2-2}
		&\textbf{Code-NN}:show me the coach flight from ci0 to ap0\\
		\cline{2-2}
		&\textbf{P-G}:first class flight from ap0 to ci0\\
		\cline{2-2}
		&\textbf{Tree2Seq}:show me the flight from ap0 to ci0 <unk> ci0 and would like the <unk> the first class fare\\
		\cline{2-2}
		&\textbf{Graph2Seq}:show me all flight from ci0 to ap0 first class\\
		\cline{2-2}
		&\textbf{T2S+CP}:what flight first class from ap0 to ci0 first class\\
		\cline{2-2}
		&\textbf{TAG}:show me the coach flight from ap0 to ci0 with first class\\
		\hline
		\multirow{8}{100pt}{\textbf{Python}:\{ i : d [ i ] for i in d if i ! = ' c ' \}}
		&CoNaLa\\
		\cline{2-2}
		&\textbf{GOLD}:remove key ' c ' from dictionary ' d '\\
		\cline{2-2}
		&\textbf{Code-NN}:remove all keys from a dictionary ' d '\\
		\cline{2-2}
		&\textbf{P-G}:select a string ' c ' in have end of a list ' d '\\
		\cline{2-2}
		&\textbf{Tree2Seq}:get a key ' key ' one ' , ' one ' , ' <unk>\\
		\cline{2-2}
		&\textbf{Graph2Seq}:filter a dictionary of dictionaries from a dictionary ' d ' where a dictionary of dictionaries ' d '\\
		\cline{2-2}
		&\textbf{T2S+CP}:find all the values in dictionary ' d ' from a dictionary ' d '\\
		\cline{2-2}
		&\textbf{TAG}:remove the key ' c ' if a dictionary ' d '\\
		\hline
	\end{tabular}

	\caption{Examples from models on different programming language related datasets}
	\label{examples}
\end{table}

%% file: sections/01_intro_3.tex
\section{Introduction}\label{sec:intro}
The comment for the programming code is critical for software development, which is crucial to the further maintenance of the project codebase with significant improvement of the readability~\citep{aggarwal2002integrated,tenny1988program}. 
Code comment generation aims to automatically transform program code into natural language with the help of deep learning technologies to boost the efficiency of the code development. 

\begin{figure}
	%Tree-to-Sequence Model
	\subfigure[Struct2Seq example]{
		\label{fig:subfig:a} %% label for first subfigure
		\includegraphics[scale=0.37]{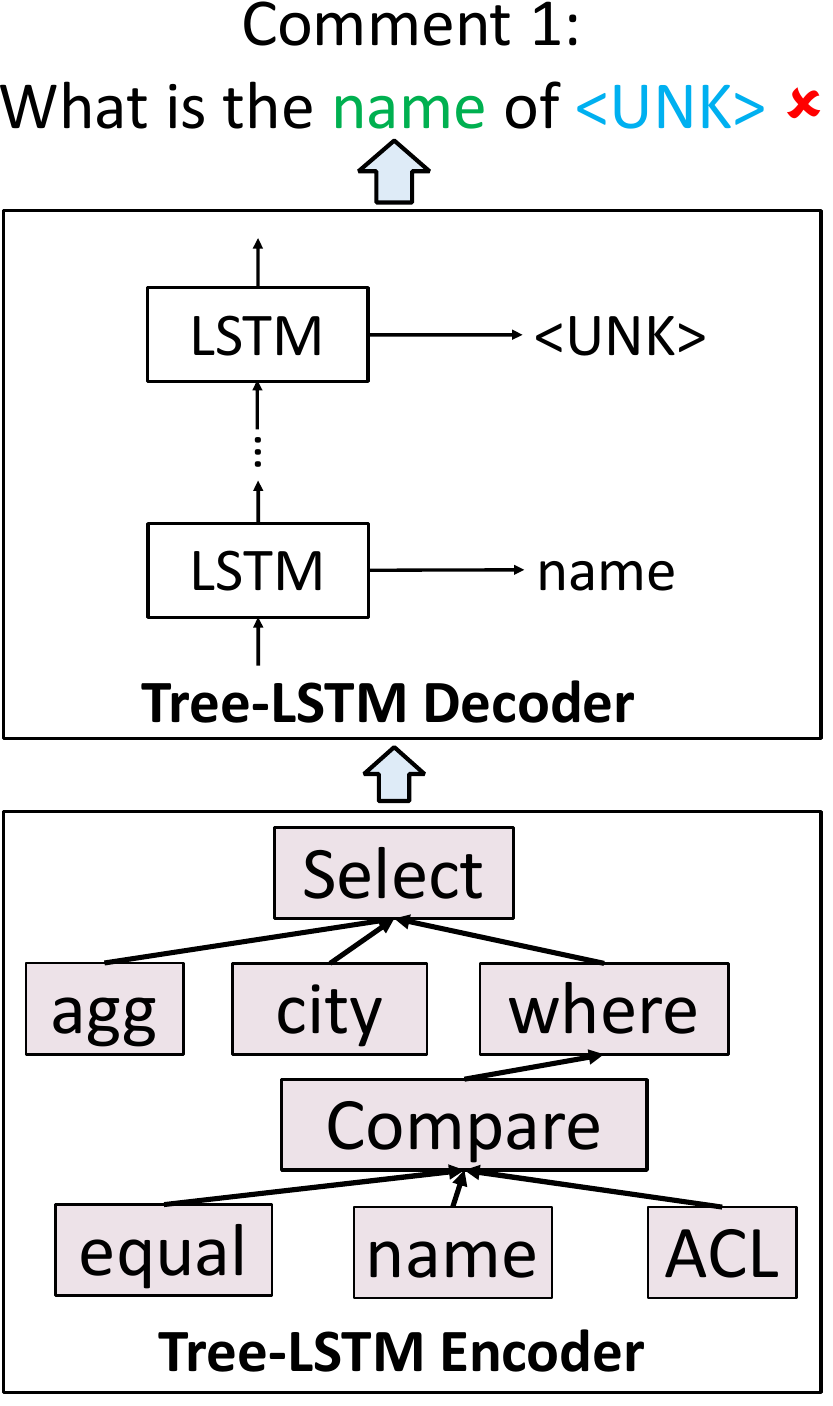}}
	\hspace{.0in}
	\subfigure[TAG example]{
		\label{fig:subfig:b} %% label for second subfigure
		\includegraphics[scale=0.37]{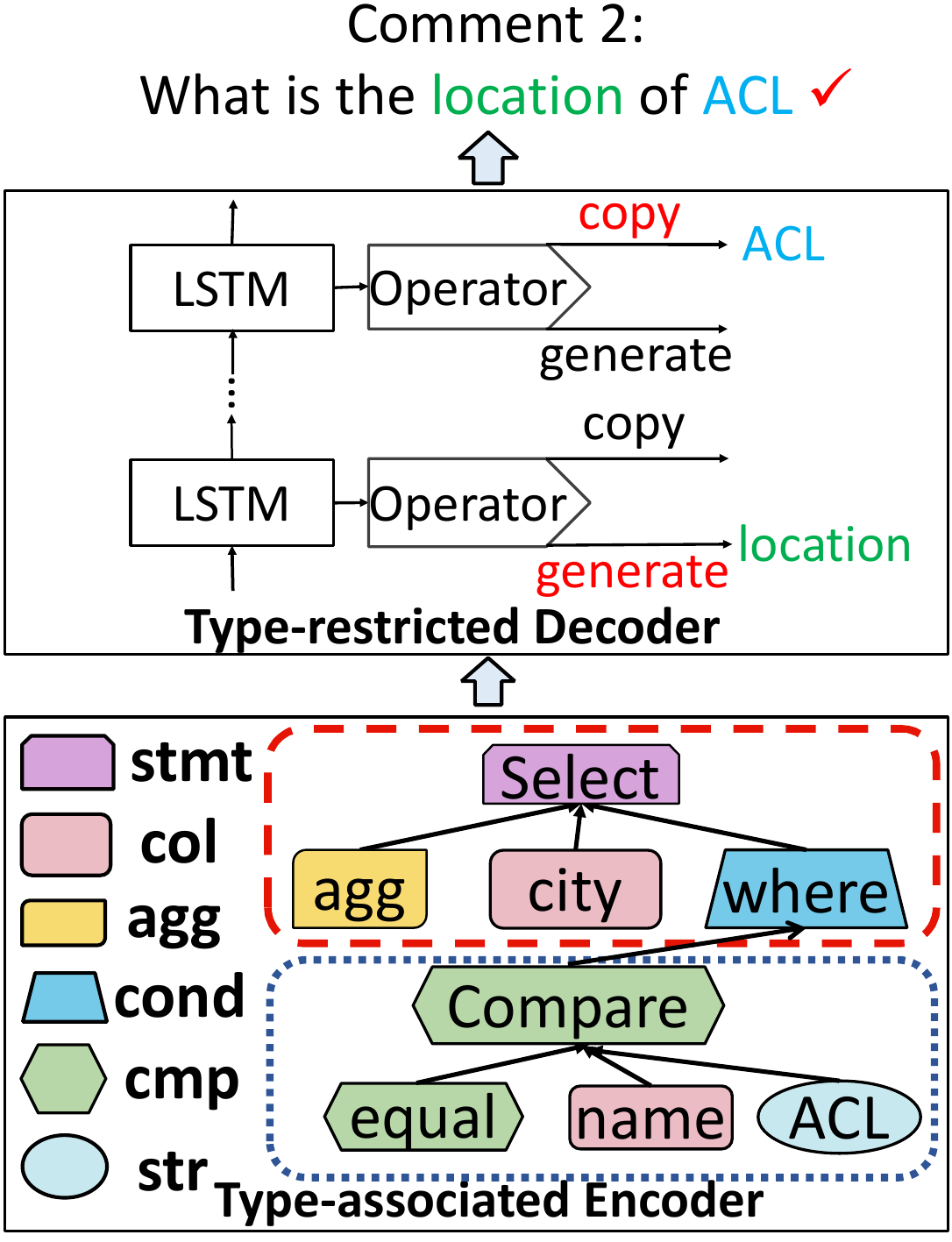}}
		\vspace{-4mm}
	\caption{Comment generation frameworks. Different types are denoted as different colors and shapes in (b).}
	\vspace{-8mm}
	\label{fig:intro} %% label for entire figure
\end{figure}

Existing leading approaches address the code comment generation task under the structure-to-sequence (Struct2Seq) framework with an encoder-decoder manner by taking advantage of the inherent structural properties of the code.
For instance, existing solutions leverage the syntactic structure of abstract syntax trees (AST) or parse trees from source code have shown significant improvement to the quality of the generated comments \citep{liang2018automatic,alon2018code2seq,hu2018deep,wan2018improving}; 
% Solutions utilizing global structural information have also shown good results \textcolor{red}{(in terms of ?)}~
Solutions representing source code as graphs have also shown high-quality comment generation abilities by taking advantage of extracting the structural information of the codes~\citep{xu-etal-2018-sql, xu2018graph2seq, fernandes2018structured}.

Although promising results were reported, we observe that the information of the node type in the code
% \st{, e.g. operator and string,} 
is not considered in these aforementioned Struct2Seq based solutions. 
The lack of such essential information lead to the following common limitations:
1) Losing the accuracy for encoding the source code with the same structure but has different types. As shown in Fig.~\ref{fig:subfig:a}, a Tree-LSTM \citep{tai-etal-2015-improved} encoder is illustrated to extract the structural information, the two subtrees of the code `Select' and `Compare' in the dashed box have the same structure but different types, with the ignorance of the type information, the traditional encoders illustrate the same set of neural network parameters to encode the tree, which leads to an inaccurate generation of the comment. 
2) Losing both the efficiency and accuracy for searching the large vocabulary in the decoding procedure, especially for the out-of-vocabulary (OOV) words that exist in the source code but not in the target dictionary. As shown in the Fig.~\ref{fig:subfig:a}, missing the type of `ACL' node usually results in an unknown word `UNK' in the generated comments. Thus, the key to tackle these limitations is efficiently utilizing the node type information in the encoder-decoder framework.

To well utilize the type information, we propose a Type Auxiliary Guiding (TAG) encoder-decoder framework.
% by introducing the type information into the N-ary tree representation of the code.
As shown in Fig.~\ref{fig:subfig:b},
in the encoding phase, we devise a \emph{Type-associated encoder} to encode the type information in the encoding of the N-ary tree.
In the decoding phase, we facilitate the generation of the comments with the help of type information in a two-stage process naming \emph{operation selection} and \emph{word selection} to reduce the searching space for the comment output and avoid the out-of-vocabulary situation.
Considering that there is no ground-truth labels for the operation selection results in the two-stage generation process, we further devised a Hierarchical Reinforcement Learning (HRL) method to resolve the
% , which apply BLEU as a reward to feedback hierarchical sampling between two stages.
training of our framework.
% 
%Contribution
Our proposed framework makes the following contributions:
\setlist{nolistsep}
\begin{itemize}[noitemsep]
	\item  An adaptive \emph{Type-associated encoder} which can summarize the information according to the node type; 
	\item  A \emph{Type-restricted decoder} with a two-stage process to reduce the search space for the code comment generation;
	\item  A hierarchical reinforcement learning approach that jointly optimizes the operation selection and word selection stages. 
\end{itemize}

\begin{figure*}[tb]
	\centering
	\includegraphics[width=2\columnwidth]{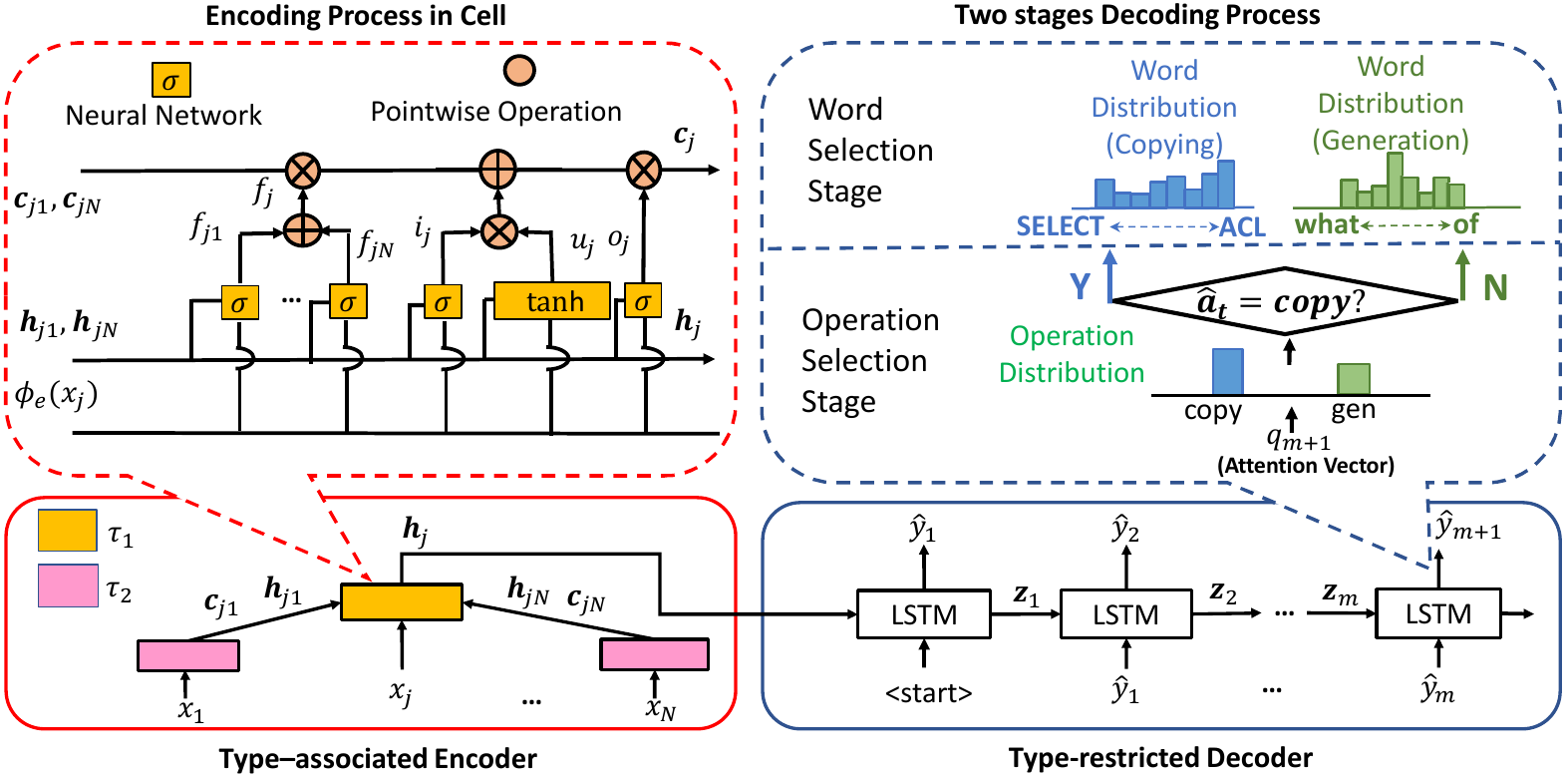}
% 	\vspace{-2mm}
	\caption{TAG Encoder and Decoder framework.}
    \vspace{-4mm}
	\label{fig:proposed_framework}
\end{figure*}

%% file: sections/02_related.tex
\section{Related Work}\label{sec:related}
%The autocommention is based str2seq.....there four 
Code comment generation frameworks generate natural language from source code snippets, e.g. SQL, lambda-calculus expression and other programming languages. 
As a specified natural language generation task, 
the mainstream approaches could be categorized into textual based method and structure-based method.

%based on token level information
The textual-based method is the most straightforward solution which only considers the sequential text information of the source code. 
For instance, \citet{movshovitz-attias-cohen-2013-natural} uses topic models and n-grams to predict comments with given source code snippets; 
\citet{iyer-etal-2016-summarizing} presents a language model Code-NN using LSTM networks with attention to generate descriptions about C\# and SQL;
\citet{allamanis2016convolutional} predicts summarization of code snippets using a convolutional attention network;
\citet{wong-mooney-2007-generation} presents a learning system to generate sentences from lambda-calculus expressions by inverting semantic parser into statistical machine translation methods.

% based on structure
The structure-based methods take the structure information into consideration and outperform the textual-based methods. 
%The other works \cite{liang2018automatic,alon2018code2seq,hu2018deep,wan2018improving} further take the tree structure into considerations, such as abstract syntax trees and parse trees. 
\citet{alon2018code2seq} processes a code snippet into the set of compositional paths in its AST and uses attention mechanism to select the relevant paths during the decoding. 
\citet{hu2018deep} presents a Neural Machine Translation based model which takes AST node sequences as input and captures the structure and semantic of Java codes.
\citet{wan2018improving} combines the syntactic level representation with lexical level representation by adopting a tree-to-sequence \citep{eriguchi-etal-2016-tree} based model. 
\citet{xu2018graph2seq} considers a SQL query as a directed graph and adopts a graph-to-sequence model to encode the global structure information.
% \st{into the node in the graph}.

% In addition, the source code can be modelled by graph. \cite{xu2018graph2seq} propose a graph-to-sequence model to encode the global structure information of a directed graph which represents the SQL query into a node.
% \textcolor{blue}{Our work also related to the research works for other natural language generation tasks, e.g. summarization, because they also utilizing copy mechanism to address OOV problem.}
Copying mechanism is utilized to address the OOV issues in the natural language generation tasks by reusing parts of the inputs instead of selecting words from the target vocabulary.
\citet{see-etal-2017-get} presents a hybrid pointer-generator network by introducing pointer network \citep{vinyals2015pointer} into a standard sequence-to-sequence (Seq2Seq) model for abstractive text summarization. 
COPYNET from \citet{gu-etal-2016-incorporating} incorporates the conventional copying mechanism into Seq2Seq model and selectively copy input segments to the output sequence.
In addition, \citet{ling-etal-2016-latent} uses the copying mechanism
% \textcolor{red}{\st{semantic parsing to copy the natural language inputs to the generated code}} 
to copy strings from the code.

%Some works also consider the case that translates natural language into the structured data, which reverse the direction, i.e. from structured data to natural language. 
Our targeted task is considered as the opposite process of natural language to programming code (NL-to-code) task. So some of the NL-to-code solutions are also taken as our references.
\citet{dong-lapata-2016-language} distinguishes types of nodes in the logical form by whether nodes have child nodes. \citet{yin-neubig-2017-syntactic,rabinovich-etal-2017-abstract,xu-etal-2018-sql} take the types of AST nodes into account and generate the corresponding programming codes. 
\citet{cai2018encoder} borrows the idea of Automata theory and considers the specific types of SQL grammar in Backus-Naur form (BNF) and generates accurate SQL queries with the help of it.

Inspired by the methods considering the type information of the code, our solution differs from the existing method with
a \emph{Type-associated Encoder} that encodes the type information during the substructure summarization and a \emph{Type-restricted Decoder} that can reduce search space for the code comment generation.
In addition, two improvements are developed according to our objectives.
First, we design a type-restricted copying mechanism to reduce the difficulty of extracting complex grammar structure from the source code. Second, 
% instead of using pseudo-labels, 
we use a hierarchical reinforcement learning methods to train the model in our framework to learn to select from either copy or other actions, the details will be presented in Section~\ref{sec:overview}.

%% file: sections/03_overview_2.tex
\section{Model Overview}\label{sec:overview}
We first make the necessary definition and formulation for the input data and the code comment generation problem for our Type Auxiliary Guiding (TAG) encoder-decoder framework.

 \begin{myDef}
	\label{type_tree}
	\textbf{Token-type-tree.} Token-type-tree $T_{x, \tau}$ represents the source code with the node set $V$, which is a rooted N-ary tree. And $V=\{v_1,v_2,..,v_{|V|}\}$ denotes a partial order nodes set satisfying $v_1\preceq v_2\preceq ...,\preceq v_{|V|}$. Let internal node $v_j=\{x_j, \tau_j\}$, where $x_j$ denotes the token sequence and $\tau_j$ denotes a type from grammar type set $\mathcal{T}$. 
\end{myDef}

Token-type-tree can be easily constructed from token information of the original source code and type information of its AST or parse tree. According to Definition \ref{type_tree}, we formulate the code comment generation task as follows.

\begin{myFor}
	\label{question}
	\textbf{Code Comment Generation with Token-type-tree as the Input.}
	Let $\mathcal{S}$ denote training dataset and labeled sample $(T_{x, \tau}, \bm{y})\in \mathcal{S}$, where $T_{x, \tau}$ is the input token-type-tree, $\bm{y}=(y_1, y_2, \cdots, y_M)$ is the ground truth comment with $M$ words. The task of code comment generation is to design a model which takes the unlabeled sample $T_{x, \tau}$ as input and predicts the output as its comment, denoted as $\bm{y}$.
\end{myFor}

% \st{Following, we will present our Type Auxiliary Guiding model (\textbf{TAG}) for the above comment generation task.}
Our framework follows the encoder-decoder manner, and consists of the revised two major components, namely the \textit{Type-associated Encoder} and \textit{Type-restricted Decoder}. 
As shown in Fig.~\ref{fig:proposed_framework}.

The \emph{Type-associated Encoder}, as shown in Fig.~\ref{fig:proposed_framework}, recursively takes the token-type-tree $T_{x,\tau}$ as input, and maintains the semantic information of the source code in the hidden states. 
Instead of using the same parameter sets to learn the whole token-type-tree, 
% \st{type-associated Tree-LSTM Cell} 
\emph{Type-associated Encoder} utilizes multiple sets of parameters to learn the different type of nodes.
The parameters of the cells are adaptively invoked according to the type of the current node during the processing of the input token-type-tree.
Such a procedure enables the structured semantic representation to contain the type information of the source code.

The \emph{Type-restricted Decoder}, as shown in the right part of Figure \ref{fig:proposed_framework}, takes the original toke-type-tree $T_{x,\tau}$ and its semantic representation from encoder as input and generates the corresponding comment. 
Different from conventional decoders which generate output only based on the target dictionary, our \emph{Type-restricted Decoder} considers both input code to the encoder and target dictionary as the source of output. Attention mechanism is employed to compute an attention vector which is used to generate the output words through a two-stage process:
(1) Determine either to copy from the original token-type-tree or to generate from the current hidden state according to the distribution of the operation.
(2) If the copying operation is selected, the words are copied from the selected node from the token-type-tree $T_{x, \tau}$ with restricted types; otherwise, the candidate word will be selected from the target dictionary.
The above two-stage process is guided by the type which is extracted from the hidden state of encoder with the help of attention mechanism. Such a process enables adaptive switching between copying and generation processes, and not only reduces the search space of the generation process but also addresses the OOV problem with the copying mechanism.

Although the proposed framework provides an efficient solution with the utilization of the type information in the code, training obstacles are raised accordingly:
(1) No training labels are provided for the operation selection stage. 
(2) There is a mismatch between the evaluation metric and the objective function. Thus, we further devised an HRL method to train our TAG model. In the HRL training, the TAG model feeds back the evaluation metric as the learning reward to train the two-stage sampling process without relying on the ground-truth label of operation selection stage.

%% file: sections/04_model_2.tex
\section{Type-associated Encoder}

% {\setlength\belowcaptionskip{-3ex}
% \begin{figure}[t]
% 	\centering
% 	\includegraphics[width=0.9\columnwidth]{./figures/encoder_cropped}
% 	\caption{Example of a SQL code portion and its tree structure.}
% % 	\textcolor{blue}{(Where is the type information in this Figure??? Need to visually express the type info in order to let the reviewer know the idea.)}}
% 	%\vspace{-4mm}
% 	\label{encoder}
% \end{figure}}

The encoder network aims to learn a semantic representation of the input source code. 
The key challenge is to provide distinct summarization for the sub-trees with the same structure but different semantics.
As shown in the Type-associated Encoder in Fig.~\ref{fig:intro}, the blue and red dashed blocks have the same 3-ary substructure.  
The sub-tree in the blue box shares the same sub-structure with the tree in the red box, which is usually falsely processed by the same cell in a vanilla Tree-LSTM.
By introducing the type information, the semantics of the two subtrees are distinguished from each other.

Our proposed Type-associated Encoder is designed as a variant $N$-ary Tree-LSTM.
Instead of directly inputting type information as features into the encoder for learning, we integrate the type information as the index of the learning parameter sets of the encoder network.
More specifically, different sets of parameters are defined through different types, which provides a more detailed summarization of the input.
As is shown in Fig.~\ref{fig:subfig:b}, the two sub-trees in our proposed Type-associated Encoder are distinguished by the type information.
% \textcolor{red}{(by how? not clear from either the figure or the description.)}.
% For instance, \textcolor{red}{feeding the tree of the SQL code portion in Fig. \ref{encoder} into the type-associated encoder, two of the blue dashed boxes share the same set of parameters, while the red dashed box uses another set of parameters.} 
% In the type-associated encoder, the parameters are not shared across the cells with different types. 
% \textcolor{green}{\st{Since the tree could contain $N$ child nodes, we assume the child nodes of these tree-structures are ordered, so the child nodes can be indexed from $1$ to $N$.}} 
The tree contains $N$ ordered child  nodes, which are indexed from $1$ to $N$.
For the $j$-th node, the hidden state and memory cell of its $k$-th child node is denoted as $\bm{h}_{jk}$ and $\bm{c}_{jk}$, respectively. 
In order to effectively capture the type information, we set $\bm{W}_{\tau_j}$ and $\bm{b}_{\tau_j}$ to be the weight and bias of the $j$-th node, and $\bm{U}_{\tau_{jk}}$ be the weight of the $k$-th child of the $j$-th node. The transition equation of the variant $N$-ary Tree-LSTM is shown as follow:

\begin{small}
\vspace{-6mm}
	\begin{equation}
	\vspace{-4mm}
	\label{input_gate}
	\bm{i}_j = \sigma\left(\bm{W}_{\tau_j}^{\left(\bm{i}\right)} \phi\left(x_j\right) + \sum_{l=1}^{N}\bm{U}_{\tau_{jl}}^{\left(\bm{i}\right)} \bm{h}_{jl} + \bm{b}_{\tau_j}^{\left(\bm{i}\right)}\right),
	\vspace{-0.6mm}
	\end{equation}
	\begin{equation}
	\bm{f}_{jk} = \sigma\left(\bm{W}_{\tau_{jk}}^{\left(\bm{f}\right)} \phi\left(x_j\right) + \sum_{l=1}^{N}\bm{U}_{\tau_{jl,k}}^{\left(\bm{f}\right)} \bm{h}_{jl} + \bm{b}_{\tau_{jk}}^{\left(\bm{f}\right)} \right), \label{forget_gate}
	\vspace{-2.0mm}
	\end{equation}
	\begin{equation}
	\bm{o}_j = \sigma\left(\bm{W}_{\tau_j}^{\left(\bm{o}\right)} \phi\left(x_j\right) + \sum_{l=1}^{N}\bm{U}_{\tau_{jl}}^{\left(\bm{o}\right)} h_{jl} + \bm{b}_{\tau_j}^{\left(\bm{o}\right)} \right),
	\vspace{-1.5mm}
	\end{equation}
	\begin{equation}
	\bm{u}_j = \tanh\left(\bm{W}_{\tau_j}^{\left(\bm{u}\right)} \phi\left(x_j\right) + \sum_{l=1}^{N}\bm{U}_{\tau_{jl}}^{\left(\bm{u}\right)} h_{jl} + \bm{b}_{\tau_j}^{\left(\bm{u}\right)}\right),
	\vspace{-2mm}
	\end{equation}
	\begin{equation}
	\bm{c}_j=\bm{i}_j \odot \bm{u}_j + \sum_{l=1}^{N}\bm{f}_{jl}\odot \bm{c}_{jl},
	\vspace{-2mm}
	\end{equation}
	\begin{equation}
	\label{hidden_state}
	\bm{h}_j = \bm{o}_j \odot \tanh\left(\bm{c}_j\right),
	\vspace{-2mm}
	\end{equation}
	\vspace{-2mm}
\end{small}
% where in Eq.~\ref{forget_gate}, $k=1,2,\cdots, N$. 

% Referring to the standard $N$-ary Tree-LSTM \citep{tai-etal-2015-improved}, 
We employ the forget gate \citep{tai-etal-2015-improved} for the Tree-LSTM, the parameters for the $k$-th child of the $j$-th node's is denoted as $\bm{f}_{jk}$. $\bm{U}_{\tau_{jl,k}}$ is used to represent the weight of the type for the $l$-th child of the $j$-th node in the $k$-th forget gate. 
The major difference between our variants and the traditional Tree-LSTM is that the parameter set ($\bm{W}_{\tau}$, $\bm{U}_{\tau}$, $\bm{b}_{\tau}$) are specified for each type $\tau$.

%% file: sections/05_typeDecoder.tex
\section{Type-restricted Decoder}
%In decoder phase, however, the NLG tasks often encounter the OOV problem, a problem that certain words of target output appear in the source token tree but not in the vocabulary. In the meanwhile, these words are usually associated with particular grammar types. Therefore, we devise the type-restricted decoder, which is shown on the right side of Figure \ref{model_framework}. This decoder phase contains two stages: operation selection stage and vocabulary selection stage. 
%The decoder network aims to generate target comments based on the representation maintained in the encoder network. However, the NLG tasks often suffer from the OOV problem. In the meanwhile, OOV words are usually associated with particular grammar types. 
%Given the $G_{\tau}$, the $G_{\tau}$  determines the type-associated Tree-LSTM cell and further forms the grammatical substructure. Therefore, after feeding $G_x$ into this encoder, hidden states are calculated recursively and finally, we can obtain the rooted hidden state.

% \st{The decoder network is responsible for generating target comments by (1) copying token from the original token-type-tree or (2) generating vocabulary from the target dictionary, which is based on the type information decoded from the current hidden states.}
% Benefiting from introducing node type in 
Following with the Type-associated Encoder, we propose a Type-restricted Decoder for the decoding phase, which incorporates the type information into its two-stage generation process.
%Specifically, we utilize attention mechanism to create attention vectors based on the similarity between the hidden states in the decoder phrase to the grammatical representation of each of the node maintained in the encoder phrase. 
First of all, an attention mechanism is adopted in the decoding phase which takes hidden states from the encoder as input and generates the attention vector.
% and maintains token and type information from token-type-tree. )} 
The resulted attention vector is used as input to the following two-stage process, named \emph{operation selection stage} and \emph{word selection stage}, respectively. 
The operation selection stage selects between generation operation and copying operation for the following word selection stage.
If the generation operation is selected, the predicted word will be generated from the targeted dictionary.
If the copying operation is selected, then a type-restricted copying mechanism is enabled to restrict the search space by masking down the illegal grammar types.
Furthermore, a copying decay strategy is illustrated to solve the issue of repetitively focusing on specific nodes caused by the attention mechanism. 
The details of each part are given below.

%After obtaining the decoder hidden state $\bm{z}_m$ at $m$-th time step from the normal LSTM decoder that is combined with the attention mechanism, we first devise an operation selection stage that decides to execute the copying or the generation. As for the generating, we simply estimate the conditional probability $p(y_m|y_1,\cdots,y_{m-1};G_x;G_{\tau})$; as for the copying, we devise a type restricted procedure that only copies the words that belong to certain types.
%\begin{figure}[ht]
%    \centering
%    \includegraphics[width=\columnwidth]{./figures/decoder}
%    \caption{A toy example of translating a SQL to a comment by utilizing the type-tree. The type in the gray blocks denotes that this types are masked, so the unmasked type can be copied directly.}
%    \label{mask}
%\end{figure}
% \subsection{Operation Selection Stage}
\textbf{Attention Mechanism:}
% First of all, we use the semantic representation extracted by the encoder, 
The encoder extracts the semantic representation as the hidden state of the rooted nodes, denoted as $\bm{h}_r$, which are used to initialize the hidden state of the decoder, $\bm{z}_0 \leftarrow \bm{h}_r$. 
At time step $m$, given output $y_{m-1}$ and the hidden state of the decoder $\bm{z}_{m-1}$ at last time step $m-1$, the hidden state $\bm{z}_m$ is recursively calculated by the LSTM cells in the decoder,
\begin{normalsize}
    %\vspace{-2mm}
	\begin{equation}
	\bm{z}_m = LSTM(\bm{z}_{m-1}, y_{m-1}).
	%\vspace{-4mm}
	\end{equation}
\end{normalsize}
The attention vector $\bm{q}$ is calculate with:
{\setlength\abovedisplayskip{1pt}
\setlength\belowdisplayskip{1pt}
\begin{normalsize}
%\vspace{-2mm}
	\begin{equation}
	\begin{split}
	\alpha_{mj} &= \frac{\exp\left(\bm{h}_j^\top \bm{z}_m\right)}{\sum_{j=1}^{|V_x|}\exp\left(\bm{h}_j^\top \bm{z}_m\right)},\\
	\widetilde{\bm{q}_m} &= \sum_{j=1}^{|V_x|}\alpha_{mj}\bm{h}_j,\\
	\bm{q}_m &= \tanh\left(\bm{W}_q\left[\widetilde{\bm{q}},\bm{z}_m\right]\right),
	\end{split}
	\end{equation} 
\end{normalsize}}
where $\bm{W}_q$ is the parameters of the attention mechanism. 
% \textcolor{red}{Such an attention mechanism extracts both the token and type information of the current state, which is further extracted in the following operation selection and vocabulary generation. (extract....extract?)} 
The attention vector contains the token and type information, which is further facilitated in the following operation selection and word selection stages.

\textbf{Operation Selection Stage:}
Operation Selection Stage determines either using the copying operation or the generation operation to select the words based on the attention vector and hidden states from the encoder.
Specifically, given the attention vector $\bm{q}_m$ at time step $m$, Operation Selection Stage estimates the conditional probabilities as the distribution of the operation $p(\hat{a}_m|\hat{y}_{<m};T_{x,\tau})$, where $\hat{a}_m \in \{0, 1\}$ and 0 and 1 represents the copy and the generation operations, respectively. 
A fully connected layer followed by a softmax is implemented to compute the distribution of the operations.
{\setlength\abovedisplayskip{1pt}
\setlength\belowdisplayskip{1pt}
\begin{normalsize}
	\begin{equation}
	\label{eq:softmx}
	p(\hat{a}_m|\hat{y}_{<m};T_{x,\tau}) = softmax(\bm{W}_s \bm{q}_m), 
	\end{equation}
\end{normalsize}}
The $\bm{W_s}$ in the Eq.~\ref{eq:softmx} is the trainable parameters. 
Since there is no ground-truth label for operation selection, we employ an HRL method to jointly train the operation selection stage and the following stage, the details are provided in Section~\ref{rl}.

\textbf{Word Selection Stage:}
Word Selection Stage also contains two branches. The selection between them is determined by the previous stage.
If the generation operation is selected in the Operation Selectoin Stage, the attention vector will be fed into a softmax layer to predict the distribution of the target word, formulated as
{\setlength\abovedisplayskip{1pt}
\setlength\belowdisplayskip{1pt}
\begin{normalsize}
	\begin{equation}
	\label{gen_branch}
	p(y_m|\hat{a}_m=1,\hat{y}_{<m};T_{x,\tau})=softmax\left(\bm{W}_g \bm{q}_m\right),
	\end{equation}
\end{normalsize}}
where $\bm{W}_g$ is the trainable parameters of the output layer.
Otherwise, if the copy operation is selected, we employ the dot-product score function to calculate score vector $\bm{s}_m$ of the hidden state of the node and the attention vector.
Similarly, score vector $\bm{s}_m$ will be fed into a softmax layer to predict the distribution of the input word, noted as:
{\setlength\abovedisplayskip{1pt}
\setlength\belowdisplayskip{1pt}
\begin{normalsize}
%	\vspace{-4mm}
	\begin{equation}
	\begin{split}
	\bm{s}_m &=\left[\bm{h}_1, \bm{h}_2, \cdots, \bm{h}_{|V_x|}\right]^\top \bm{q}_m \\
	p(y_m|\hat{a}_m&=0;\hat{y}_{<m};T_{x,\tau})=softmax\left(\bm{s}_m\right).
	\end{split}
	\end{equation}
%	\vspace{-4mm}
\end{normalsize}}
% \textcolor{red}{(Then, how to do the selection of the word?)}
One step further, to filter out the illegally copied candidates, we involve a grammar-type based mask vector $\bm{d}_m \in \mathbb{R}^{|V_x|}$ at each decoding step $m$. Each dimension of $\bm{d}_m$ corresponds to each node of the token-type-tree.
If the mask of the node in token-type-tree indicates the node should be filtered out, then the corresponding dimension is set as negative infinite. Otherwise, it is set to $0$. Thus, the restricted copying stage is formulated as
{\setlength\abovedisplayskip{1pt}
\setlength\belowdisplayskip{1pt}
\begin{normalsize}
%\vspace{-2mm}
	\begin{equation}
	\label{no_copy_one}
	p(y_m|\hat{a}_m=0,\hat{y}_{<m};T_{x,\tau})=softmax\left(\bm{s}_m + \bm{d}_m\right).
	\end{equation}
%\vspace{-4mm}
\end{normalsize}}
The word distribution of the two branches is represented with a softmax over input words or target dictionary words in Eq. \ref{gen_branch} and Eq. \ref{no_copy_one}. At each time step, the word with the highest probability in the word distribution will be selected.

\textbf{Copying Decay Strategy:}
Similar to the conventional copying mechanism, we also use the attention vector as a pointer to guide the copying process. 
The type-restricted copying mechanism tends to pay more attention to specific nodes, resulting in the ignorance of other available nodes, which makes certain copied tokens repeatedly active in a short distance in a single generated text, lead to a great redundancy of the content.
	
So we design a \emph{Copying Decay Strategy} to smoothly penalize certain probabilities of outstandingly copied nodes. We define a copy time-based decay rate $\lambda_{mi}$ for the $i$-th tree node $x_{i}$ in the $m$-th decoding step.
% \textcolor{red}{\st{The decay rate of $\lambda_{mi}$ is set as a coefficient $\gamma \in (0, 1)$ to control the amount of the preserved decay rate.}} 
% A decay rate of $\lambda_{mi}$ is set to control the amount of the preserved decay rate.}
%The decay rate is updated in the following two steps: 
If one node is copied in time step $m$, its decay rate is initialized as $1$. 
In the next time step $m+1$, it is scaled by a coefficient $\gamma \in (0, 1)$: 
{\setlength\abovedisplayskip{1pt}
\setlength\belowdisplayskip{1pt}
\begin{equation}
\begin{aligned}
\lambda_{m+1, i} = \gamma\lambda_{m,i}
\end{aligned}
\end{equation}}
The overall formulation for the Type-restricted Decoder is:
{\setlength\abovedisplayskip{1pt}
\setlength\belowdisplayskip{1pt}
\begin{normalsize}
	\begin{equation}
	\begin{aligned}
	p(y_m|&\hat{a}_m=0,\hat{y}_{<m};T_{x,\tau})=\\
	&softmax\left(\bm{s}_m + \bm{d}_m\right)\odot (1-\bm{\lambda}_m)
	\end{aligned}
	%\vspace{-4mm}
	\end{equation}
\end{normalsize}}

%% file: sections/06_rl.tex
\section{Hierarchical Reinforcement Learning}\label{rl}
% \st{In this section, we will delve into the training strategy for the aforementioned encoder-decoder framework.} 
% \st{We present a new encoder-decoder framework by introducing type information guiding comment generation process.} 
There remain two challenges to train our proposed framework, which are 1) the lack of ground truth label for the operation selection stage and 2) the mismatch between the evaluation metric and objective function.
Although it is possible to train our framework by using the maximum likelihood estimation (MLE) method which constructs pseudo-labels or marginalize all the operations in the operation selection stage \citep{jia-liang-2016-data, gu-etal-2016-incorporating}, 
%However, when output words appear in both the target dictionary and source code input, the construction guidelines for pseudo labels generally tend to \textcolor{blue}{copy or generation, which leads to model selection preferences}.
%Some recent works also discover that the MLE method leads to a mismatch between evaluation metric and the objective function~\cite{keneshloo2019deep, ranzato2015sequence}.
the loss-evaluation mismatch between MLE loss for training and non-differentiable evaluation metrics for testing lead to inconsistent results~\cite{keneshloo2019deep, ranzato2015sequence}.
% \st{For comment generation tasks, mainstream methods generally use BLEU or Rouge as evaluation metric, which can not use for model training.} 
To address these issues, we propose a Hierarchical Reinforcement Learning method to train the operation selection stage and word selection stage jointly.

We set the objective of the HRL as maximizing the expectation of the reward $R(\hat{\bm{y}}, \bm{y})$ between the predicted sequence $\hat{\bm{y}}$ and the ground-truth sequence $\bm{y}$, denoted as $L_{r}$.
It could be formulated as a function of the input tuple $\{T_{x, \tau}, \bm{y}\}$ as,
{\setlength\abovedisplayskip{1pt}
\setlength\belowdisplayskip{1pt}
% \begin{small}
%\vspace{-4mm}
	\begin{equation}
	\begin{split}
	\label{expecation}L_{r}&=\frac{1}{|\mathcal{S}|}\sum_{(T_{x,\tau}, \bm{y}) \in \mathcal{S}}E_{\hat{\bm{y}} \sim p(\hat{\bm{y}}|T_{x,\tau})}[R(\hat{\bm{y}}, \bm{y})]\\
	&=\frac{1}{|\mathcal{S}|}\sum_{(T_{x,\tau}, \bm{y}) \in \mathcal{S}}\sum_{\hat{\bm{y}} \in Y} p(\hat{\bm{y}}|T_{x,\tau})R(\hat{\bm{y}}, \bm{y}),
	\end{split}
%	\vspace{-4mm}
	\end{equation}}
% \end{small}
Here, $Y$ is the set of the candidate comment sequences. The reward $R(\hat(\bm{y}), \bm{y})$ is the non-differentiable evaluation metric, i.e., BLEU and ROUGE (details are in Section~\ref{sec:eva}). 
The expectation in Eq. \eqref{expecation} is approximated via sampling $\hat{\bm{y}}$ from the distribution $p(\hat{\bm{y}}|T_{x,\tau})$. 
The procedure of sampling $\hat{\bm{y}}$ from $p(\bm{\hat{y}}|T_{x,\tau})$ is composed of the sub-procedures of sampling $\hat{y}_m$ from $p(\hat{y}_m|\hat{y}_{<m};T_{x,\tau})$ in each decoding step $m$.
%$R(\hat{\bm{y}}, \bm{y})$ represents the evaluation measurement.
% , in our case, either BLEU \cite{papineni-etal-2002-bleu} or ROUGE~\cite{rouge}, which 
%However, it is a sentence level reward and unable to be obtained until the whole predicted sequence $\hat{\bm{y}}$ is generated.

%After involving the HRL, Eq. $\eqref{expecation}$ can be derived with the %\textcolor{red}{two-stage joint distribution}:
As mentioned above, the predicted sequence $\hat{\bm{y}}$ comes from the two branches of Word Selection Stage, depending on the Operation Selection Stage. $a$ is defined as the action of the Operation selection stage. After involving the action $a_m$ in time step $m$, Eq. $\eqref{expecation}$ can be constructed by the joint distribution of the two stages:

\begin{small}
\begin{equation}
\setlength{\abovedisplayskip}{0.01pt}
\begin{aligned}
&\frac{1}{|\mathcal{S}|}\sum_{(T_{x,\tau}, \bm{y}) \in \mathcal{S}}\label{longtermexpectation}\sum_{\bm{\hat{y}} \in Y} p(\bm{\hat{y}}|T_{x,\tau}) R(\bm{\hat{y}}, \bm{y})\\
% &=\frac{1}{|\mathcal{T}|}\sum_{(T_{x,\tau}, \bm{y}) \in \mathcal{T}}\sum_{\hat{\bm{y}} \in Y} (\prod_{m=1}^{M} p(\hat{y}_m|\hat{y}_{<m}; T_{x,\tau})) R(\hat{\bm{y}}, \bm{y})\\
&=\frac{1}{|\mathcal{S}|}\sum_{...}\sum_{\bm{\hat{y}} \in Y}(\prod_{m=1}^{M} \sum_{\hat{a}_m}\underbrace{p(\hat{y}_m, \hat{a}_m|\hat{y}_{<m}; T_{x,\tau})}_{\text{Two-stage Joint Distribution}}) R(\bm{\hat{y}}, \bm{y})\\
&=...\underbrace{p(\hat{y}_m\!|\hat{a}_m;\!\hat{y}_{<m};\!T_{x,\tau})}_{\text{Word Distribution}}\underbrace{p(\!\hat{a}_m\!|\hat{y}_{<m}\!;\!T_{x,\tau})}_{\text{Operation Distribution}}...\\ 
\end{aligned}
%\vspace{-4mm}
\setlength{\belowdisplayskip}{0.01pt}
\end{equation}
\end{small}

%Furthermore, the two-stage joint distribution decomposes as follow:
%\begin{small}
%	\begin{equation}\label{joint_distribution}
%	\begin{aligned}
% p(\hat{y}_m, \hat{a}_t|&\hat{y}_{<m}; T_{x,\tau})=\\ &\underbrace{p(\hat{y}_m\!|\hat{a}_m;\!\hat{y}_{<m};\!T_{x,\tau})}_{\text{Vocabulary Distribution}}\underbrace{p(\!\hat{a}_m\!|\hat{y}_{<m}\!;\!T_{x,\tau})}_{\text{Operation Distribution}}.\\ 
%	\end{aligned}
%	\end{equation}
%\end{small}

% \textcolor{red}{\st{The two-stage joint distribution, which are the operation distribution $p(\hat{a}_m|\hat{y}_{<m}; T_{x,\tau})$ and the word distribution $p(\hat{y}_m|\hat{a}_m;\hat{y}_{<m};T_{x,\tau})$, where the $\hat{y}_m$ sampled from $p(\hat{y}_m|\hat{a}_m;\hat{y}_{<m};T_{x,\tau})$ is depended on $\hat{a}_m$ sampled from $p(\hat{a}_m|\hat{y}_{<m}; T_{x,\tau})$, constructing the hierarchical dependence.}}

As shown in Eq. \eqref{longtermexpectation}, 
% \st{the two-stage joint distribution $p(\hat{y}_m,\hat{a}_m|\hat{y}_{<m};T_{x,\tau})$ can be decomposed into the operation distribution $p(\hat{a}_m|\hat{y}_{<m}; T_{x,\tau})$ and the word distribution $p(\hat{y}_m|\hat{a}_m;\hat{y}_{<m};T_{x,\tau})$, which are modeled in the operation selection stage and the word selection stage respectively.} 
the model finally selects the word $\hat{y}_m$ in time step $m$ from the word distribution conditioned on $\hat{y}_{<m}$, $T_{x, \tau}$ and the operation $\hat{a}_m$ which is determined in the operation selection stage. In other words, there is a hierarchical dependency between the word selection stage and the operation selection stage.
%Equation \eqref{longtermexpectation} indicates that we can sample $\hat{y}_m$ and $\hat{a}_m$ from $p(\hat{y}_m|\hat{a}_m,\hat{y}_{<m};T_{x,\tau})$ and $p(\hat{a}_m|\hat{y}_{<m}; T_{x,\tau})$ respectively in the same decoding step $m$
%\textcolor{red}{\st{through the maximum iteration step $M$, leading to the objective as maximizing}}
% \st{Consistent with Eq. \eqref{expecation}, Eq. \eqref{longtermexpectation} approximates the expectation via sampling $\hat{y}_m$ and $\hat{a}_m$ from $p(\hat{y}_m|\hat{a}_m,\hat{y}_{<m};T_{x,\tau})$ and $p(\hat{a}_m|\hat{y}_{<m}; T_{x,\tau})$ respectively in the same decoding step $m$.}

As mentioned above, $Y$ represents the space for all candidate comments, which is too large to practically maximize $L_r$. 
Since decoding is constructed via sampling from $p(\hat{y}_m|\hat{a}_m, \hat{y}_{<m}; T_{x,\tau})$ and $p(\hat{a}_m|\hat{y}_{<m}; T_{x,\tau})$,
We adopt the Gumbel-Max solution \citep{gumbel1954statistical} for the following sampling procedure:
{\setlength\abovedisplayskip{1pt}
\setlength\belowdisplayskip{1pt}

\begin{equation}
\begin{split}
\label{sampling}
&\hat{a}_m \sim p(\hat{a}_m|\hat{y}_{<m};T_{x,\tau}),\\
&\hat{y}_m \sim p(\hat{y}_m|\hat{a}_m, \hat{y}_{<m};T_{x,\tau}).
\end{split}
\end{equation}}

Through the maximum sampling step M, Eq. (\ref{longtermexpectation}) could be further approximated as the following equation:
{\setlength\abovedisplayskip{1pt}
\setlength\belowdisplayskip{1pt}
% \begin{small}
\begin{equation}
\begin{split}
\label{reinforceobj}
\hat{L}_{r}&=\frac{1}{|\mathcal{S}|}\sum_{\bm{y} \in \mathcal{S}}R(\hat{\bm{y}},\bm{y}) 
%&\hat{y}_m \sim p(\hat{y}_m|\hat{a}_m, \hat{y}_{<m};T_{x,\tau}),\\
%&\hat{a}_m \sim p(\hat{a}_m|\hat{y}_{<m};T_{x,\tau}).
\end{split}
\end{equation}}
% \end{small}
%Based on the objective function above, in each decoding step $m$, the decoding procedure is constructed by hierarchical sampling of $p(\hat{y}_m|\hat{a}_m, \hat{y}_{<m}; T_{x,\tau})$ and $p(\hat{a}_m|\hat{y}_{<m}; T_{x,\tau})$. 
%We adopt the Gumbel-Max solution \citep{gumbel1954statistical} for the above sampling procedure.

The objective in Eq. \eqref{reinforceobj} remains another challenge: for the entire sequence $\hat{\bm{y}}$, there is only a final reward $R(\hat{\bm{y}}, \bm{y})$ available for model training, 
% \textcolor{blue}{\st{It means that our model samples tremendous amount of $\hat{y}_m$ and $\hat{a}_m$ from the large output space, but only one terminal reward $R(\hat{\bm{y}}, \bm{y})$ is available from them,} 
which is a sparse reward and leads to inefficient training of the model. 
So we introduce reward shaping \citep{ng1999policy} strategy to provide intermediate rewards to proceed towards the training goal, which adopts the accumulation of the intermediate rewards to update the model.

% \st{Instead of using the terminal reward $R(\hat{\bm{y}}, \bm{y})$ as \st{intermediate rewards across all the decoding steps} without any distinction, the cumulative reward is required. 
% Note that let $r_m(\hat{y}_m, \bm{y})=R(\hat{\bm{y}}_{1...m}, \bm{y})-R(\hat{\bm{y}}_{1...m-1}, \bm{y})$, our intermediate reward in the decoding step $m$ is computed by the cumulative reward of $r_m(\hat{y}_m, \bm{y})$,  i.e. $\sum_{\epsilon=m}^{M}r_{\epsilon}(\hat{y}_{\epsilon}, \bm{y})$.
% And then the cumulative reward plays role in updating the model in place of the terminal reward.}
% \st{It is demonstrated that using that shaped reward $r_m$ instead of awarding the terminal reward $R(\hat{\bm{y}}, \bm{y})$ does not change the optimal policy} \cite{ng1999policy}.
%
% Instead of using terminal reward, we use cumulative reward for the model update to avoid the lack of distinction of them.

To further stabilize the HRL training process, we combine our HRL objective with the maximum-likelihood estimation(MLE) function according to ~\citet{wu-etal-2018-study,wu2016google,li-etal-2017-adversarial,wu2018adversarial}:
{\setlength\abovedisplayskip{1pt}
\setlength\belowdisplayskip{1pt}
% \begin{small}
%\vspace{-2mm}
\begin{equation}
\begin{split}
L_{e}&=\frac{1}{|\mathcal{S}|}\sum_{(T_{x,\tau}, \bm{y}) \in \mathcal{S}}\sum_{\hat{\bm{y}} \in Y} {\rm log} p(\bm{y}|T_{x,\tau})\\
L&=\mu L_{e} + (1-\mu) \hat{L}_{r},
\end{split}
% \vspace{-2mm}
\end{equation}}
% \end{small}
where $\mu$ is a variational controlling factor that controls the trade-off between maximum-likelihood estimation function and our HRL objective. In the current training step $tr$, $\mu$ varies according to the training step $tt$ as follows:
{\setlength\abovedisplayskip{1pt}
\setlength\belowdisplayskip{1pt}
%\vspace{-4mm}
\begin{equation}
\mu = 1 - \frac{tr}{tt}
% \vspace{-2mm}
\end{equation}}

%% file: sections/07_exp.tex
%2. type决定子结构，子结构在encoder的作用：所有树是由若干个子结构（有限）组成，type使得子结构被发现， 结构相同，value不同:
%1. type tree有很多相同的子结构组成
%4. 不同的子结构使用同一套参数，问题：不能捕捉子结构
%3. 相同的子结构使用同一套参数，不同的子结构使用不同参数，为什么这样做

%引入type之后发现，所有树可以看作由若干个子结构组成，如例所示，
%不同的节点值有相同的子结构，意味着这些点的几何关系相同;
%传统的Tree-LSTM方法，没有使用type(没有type的引导)，不能显式地使用子结构，better representaion->相同的子结构用同一组参数

\section{Evaluation and Analysis}\label{sec:eva}

\subsection{Experimental Setup}

\subsubsection{Datasets}
We evaluate our TAG framework on three widely used benchmark data sets, which are WikiSQL \citep{zhong2017seq2sql}, ATIS \citep{dong-lapata-2016-language} and CoNaLa \citep{yin2018learning}. 
WikiSQL is a dataset of 80654 hand-annotated examples of SQL query and natural language comment pairs distributed across 24241 tables from Wikipedia. These SQL queries are further split into training (56355 examples), development (8421 examples) and test (15878 examples) sets. 
ATIS is in the form of lambda-calculus, which is a set of 5410 inquiries for flight information containing 4434 training examples, 491 development examples and 448 test examples. 
CoNaLa is a python related dataset. Its original version is used which includes 2879 snippet/intent pairs crawled from Stack Overflow, split into 2379 training and 500 test examples. We extract 200 random examples from its training set as the development set.

% To evaluate the performance of our TAG framework with different programming languages, the above three data sets are further transformed into tree structured data. 
We transfer the SQL queries of WikiSQL into ASTs with 6 types according to the Abstract Syntax Description Language (ASDL) grammar, where the ASDL grammar for SQL queries is proposed in \citet{yin-neubig-2017-syntactic}. We transfer the lambda-calculus logical forms of ATIS to tree structure with 7 types according to the method proposed in \citet{dong-lapata-2016-language}. 
The python snippets of CoNaLa are transformed into ASTs with 20 types, following the official ASDL grammar of python\footnote{https://docs.python.org/3.5/library/ast.html}. The data of the ASTs of these datasets is shown in Table \ref{statistics}, where the maximum depth of ASTs (Max-Tree-Depth), the maximum number of child nodes in ASTs (Max-Child-Count) and the average number of tree nodes in ASTs (Avg-Tree-Node-Count) are shown. 
% We assume that various source code is grammar-structured, which ensures the source code can be represented using AST trees. We believe such assumption is universally satisfied in the existing programming languages.

%Table \ref{tab6} shows the types and available types in the data set. We can find that the type contains rich information of the code. For example, there are types in column ``Types" of  Table \ref{tab6} for a abstract syntax tree of a SQL query on WikiSQL. And type ``column\_name" and type ``string", which are in column ``Avail. Types in Dec.", are allowed to be considered in Type-Restricted Decoder.

\input{sections/statistics}

%Appendix A shows the types and available types over all the data sets.
\begin{table*}[htb]
	\centering
	\normalsize
	\resizebox{\textwidth}{!}{
	\begin{tabular}{p{55pt}<{\centering}|p{38pt}<{\centering}p{47pt}<{\centering}p{48pt}<{\centering}|p{38pt}<{\centering}p{47pt}<{\centering}p{48pt}<{\centering}|p{38pt}<{\centering}p{47pt}<{\centering}p{48pt}<{\centering}}
		\toprule[1.2pt]
		\multirow{2}{*}{Model}&
		\multicolumn{3}{c|}{WikiSQL (SQL)}&
		\multicolumn{3}{c|}{ATIS (lambda-calculus)}&
		\multicolumn{3}{c}{CoNaLa (Python)}\\
		\cline{2-10}
		&BLEU-4&ROUGE-2&ROUGE-L&BLEU-4&ROUGE-2&ROUGE-L&BLEU-4&ROUGE-2&ROUGE-L\\
		\midrule[1.2pt]
		Code-NN&6.7&9.7&30.9&37.1&43.28&59.4&8.1&12.2&26.1\\
		P-G&25.7&29.2&50.1&41.9&47.3&60.5&10.0&13.8&28.0\\
		Tree2Seq&22.0&22.0&43.4&40.1&47.2&60.9&6.6&9.2&25.2\\ % \cite{eriguchi2016tree}
		Graph2Seq &17.6&24.3&45.7&34.6&41.8&58.3&10.4&14.1&28.2\\ %\cite{xu2018graph2seq}
		T2S+CP&31.0&36.8&54.5&39.0&43.7&58.4&13.3&18.5&31.5\\
		\midrule[1.2pt]
		\textbf{TAG(B)}&\textbf{35.8}&41.0&57.8&\textbf{42.4}&\textbf{47.4}&61.2&\textbf{14.1}&19.4&31.8\\
		\textbf{TAG(R)}&35.2&\textbf{41.1}&\textbf{58.1}&40.6&47.1&\textbf{61.5}&12.6&\textbf{19.7}&\textbf{32.2}\\
		\bottomrule[1.2pt]
	\end{tabular}}
% 	\vspace{-2mm}
	\caption{Comparisons with baseline models on different test sets.}
% 	\vspace{-4mm}
	\label{main_experiment}
\end{table*}

\subsubsection{Baselines Frameworks}
%For AST-form experiments, we choose the attention based sequence to sequence (\emph{Seq2Seq}) model proposed by \cite{luong-etal-2015-effective} as a baseline. 
We choose the representative designs for code comment generation as our baselines for comparison.
Code-NN \citep{iyer-etal-2016-summarizing} is chosen because of it is the first model to transform the source code into sentences.
Pointer Generator \citep{see-etal-2017-get} (P-G) is a seq2seq based model with a standard copying mechanism.
In addition, we choose the attention based Tree-to-Sequence (Tree2Seq) model proposed by \citet{eriguchi-etal-2016-tree}. 
Moreover, we also add the copying mechanism into Tree2Seq model as another baseline (T2S+CP). 
%\st{which is provided with the pseudo labels in the operation selection stage according to the existence of the ground-truth word in the tree.}
We choose Graph-to-Sequence (Graph2Seq) \citep{xu2018graph2seq} as a graph-based baseline for comparison. Since the authors have not released the code for data-preprocessing, we convert the tree-structured representation for the source code of SQL data into directed graphs for our replication.
% The results are based on the original implementations of baselines.\textcolor{blue}{(Then what is the purpose of our change in the aforementioned blue words?)}

%We train these baselines via maxinum-likelihood estimation as normal. %With the consideration that types can be viewed as a feature of a tree node, it is reasonable and straightforward to input a type as a feature embedding. So we also run a baseline \emph{Tree2Seq + Type Feat} which concatenates the type embedding and the value embedding and feeds them into the classical Tree2Seq model. It is noted that we train these baselines via maxinum-likelihood estimation as normal except for \emph{Tree2Seq + Type Feat} that is trained via reinforcement learning. %And for lambda-tree-form experiments which aims to evaluate the effectiveness of considering type of tree nodes following other type definitions compared with the traditional tree-based baselines, we abandon \emph{Seq2Seq}. 

\subsubsection{Hyperparameters} 
%\textcolor{blue}{(Is the batch size matters in the hyperparameter???)}
Code-NN uses embedding size and hidden size both as 400, and applies random uniform initializer with 0.35 initialized weight, and adopts stochastic gradient descent algorithm to train the model with a learning rate at 0.5. 
P-G uses 128 embedding size, 256 hidden size and applies random uniform initializer with 0.02 initialized weights for initialization and Adam optimizer to train the model with 0.001 learning rate. 
Graph2Seq uses 100 embedding size, 200 hidden size and applies the truncated normal initializer for initialization. Adam optimizer is used to train the model with a 0.001 learning rate.

We use the Xavier initializer \citep{glorot2010understanding} to initialize the parameters of our proposed TAG framework. The size of embeddings is equivalent to the dimensions of LSTM states and hidden layers, which is 64 for ATIS and CoNaLa and 128 for WikiSQL. 
TAG is trained using the Adam optimizer \citep{kingma2014adam} with a learning rate of 0.001. 
In order to reduce the size of the vocabulary, low-frequency words are not kept in both the vocabulary for the source codes and the vocabulary for target comments. Specifically, the minimum threshold frequency for WikiSQL and ATIS is set as 4 while for CoNaLa it is set as 2.
The hyperparameters of Tree2Seq and T2S+CP is equivalent to ours. 
The minibatch size of all the baseline models and ours are set to 32.

\subsubsection{Evaluation Metric}
We illustrate the n-gram based BLEU \citep{papineni-etal-2002-bleu} and ROUGE \citep{lin2004rouge} evaluations to evaluate the quality of our generated comments and also use them to set the reward in the HRL based training.
Specifically, BLEU-4, ROUGE-2 and ROUGE-L are used to evaluate the performance of our model since they are the most representative evaluation metric for context-based text generation.

\subsection{Results and Analysis}
\subsubsection{Comparison with the Baselines}

Table \ref{main_experiment} presents the evaluation results of the baseline frameworks and our proposed ones. 
Since our HRL could be switched to different reward functions, we evaluate both the BLEU oriented and ROUGE oriented training of our framework, denoted as TAG(B) and TAG(R).
The results of TAG(B) and TAG(R) varies slightly compared to each other.
However, both of them are significantly higher than all the selected counterparts, which demonstrates the state-of-the-art generation quality of our framework on all the datasets with different programming languages. 

Specifically, TAG improves over 15\% of BLEU-4, over 10\% of ROUGE-2 and 6\% of ROUGE-L on WikiSQL when compared to T2S+CP, which is the best one among all the baseline target for all the evaluations.
For the lambda-calculus related corpus, TAG improves 1.0\% of BLEU, 0.2\% ROUGE-2 and 0.5\% ROUGE-L on ATIS. 
The performance is more difficult to be improved on ATIS than the other two corpora due to the great dissimilarity of sub-trees of the lambda-calculus logical forms in it.
In terms of the python related corpus, TAG improves 6\% of BLEU, 6.4\% of ROUGE-2 and 2.2\% of ROUGE-L on CoNaLa when compared to the best one in our baselines. 
The low evaluation score and improvement of CoNaLa are due to the complex grammatical structures and lack of sufficient training samples, i.e., 20 types across only 2174 training samples, which result in an inadequately use of the advantage of our approach.
However, our TAG framework still outperforms all the counterparts on these two datasets.
\begin{table}[!]
	\centering
	\small
	\resizebox{0.48\textwidth}{!}{
	\begin{tabular}{p{44pt}<{\centering}|p{44pt}<{\centering}|p{44pt}<{\centering}|p{44pt}<{\centering}}
		\toprule[1.2pt]
		Model&BLEU-4&ROUGE-2&ROUGE-L\\
		\midrule[1.2pt]
		TAG-TA&34.8(-1.4)&41.0(-1.3)&57.8(-1.6)\\
		TAG-MV&35.2(-1.0)&41.1(-1.2)&58.1(-1.3)\\
		TAG-CD&33.5(-2.7)&40.0(-2.3)&57.1(-2.3)\\
		TAG-RL&34.6(-1.6)&41.4(-0.9)&58.7(-0.7)\\
		\midrule[1.2pt]
		TAG(B)&\textbf{36.2}&42.0&58.8\\
		TAG(R)&35.6&\textbf{42.3}&\textbf{59.4}\\
		\bottomrule[1.2pt]
	\end{tabular}}
	\caption{Ablation study of TAG framework.}
    \vspace{-4mm}
	\label{ablation_study}
\end{table}

\input{sections/case_study}

\subsubsection{Ablation Study}
To investigate the performance of each component in our model, we conduct ablation studies on the development sets. Since all the trends are the same, we omit the results on the other data sets and only present the ones of WikiSQL. 
The variants of our model are as follows:
\setlist{nolistsep}
\begin{itemize}[noitemsep]
	\item TAG-TA: remove \emph{Type-associated Encoder}, use Tree-LSTM instead.
	\item TAG-MV: remove the mask vector $\bm{d}_m$.
	%\item TAG-NS: we only remove the node stack $\beta_m$ (Eq. \ref{stack}) in each decoding step $m$.
	\item TAG-CD: remove Copying Decay Strategy.
%	\item TAG-NETR: we use the classical Tree-LSTM as the encoder and Type-Restricted Decoder with the concatenation of type embedding and value embedding as input.
	\item TAG-RL replace HRL with MLE,
	%\st{to determine the probability of the selected action } 
	marginalize the actions of the operation selection.

\end{itemize}

The results of the ablation study are given in Table \ref{ablation_study}. 
Overall, all the components are necessary to TAG framework and providing important contributions to the final output. %From the table, we can find that the type-associated encoder is the most important component,  
%\textbf{Stduy on Type Information} 
%As shown in Table \ref{tab3} and \ref{tab1}, we observe that the performance of TAG-NETR, that simply concatenate the token feature and the type feature and feed it into the Tree-LSTM model, is lower that standard TAG but it is still better than all other comparision baselines. This experiment result prove that the type information is profitable in our task.
% \textit{Stduy on Type-Associated Encoder}: 
When compared to TAG-TA, the high performance of standard TAG benefits from the \emph{Type-associated Encoder} which adaptively processes the nodes with different types and extracts a better summarization of the source code.
% However, the performance is lower than that of Tree2Seq and Tree2Seq+Copy, this is because we have destroy our integrated TAG architecture and when we remove the type-associated encoder, we can not obtain the type auxiliary representation, so it further leads to the drop of the performance.
% \textit{Study on Type-Restricted Decoder}:
The downgraded performance of TAG-MV and TAG-CD indicates the advantages of the type-restricted masking vector and Copying Decay Strategy. These together ensure the accurate execution of the copy and word selection.
% also play an important role in word selection stage and promote the copying mechanism to perform more accurately and more smoothly. 
% Only the \emph{Type-associated Encoder} cannot perform a good final result.
% \textit{Study of Hierarchical RL:}
The comparison of TAG and TAG-RL shows the necessity of the HRL for the training of our framework. 
% Moreover, TAG(B) and TAG(R) are trained with different reward objectives outline the advantage of the match of evaluation metric and training reward selection.

\subsubsection{Case Study}
In order to show the effectiveness of our framework in a more obvious way, some cases generated by TAG are shown in Table \ref{case_study}. SQL and Python are taken as the targeted programming languages. The comments generated by TAG show great improvements when compared to the baselines. 
Specifically, for the case in SQL, the keyword ``Otkrytie Area" is missing in all the baselines but accurately generated by our framework. 
For the case in Python, the comment generated by TAG is more readable than the others.
These cases demonstrate the high quality of the comments generated by our TAG framework.

%% file: sections/statistics.tex
\begin{table}[!htb]
	\centering
	\small
	\resizebox{0.48\textwidth}{!}{
	\begin{tabular}{p{80pt}<{\centering}|p{32pt}<{\centering}|p{32pt}<{\centering}|p{32pt}<{\centering}}
		\toprule[1.2pt]
		Dataset&WikiSQL&ATIS&CoNaLa\\
		\midrule[1.2pt]
		Max Tree Depth&5&18&28\\
		\cline{1-4}
		Max Child Num&4&15&10\\
		\cline{1-4}
		Avg Tree Node Count&11.11&33.54&28.37\\
		\bottomrule[1.2pt]
	\end{tabular}}
% 	\vspace{-2mm}
	\caption{Statistics of ASTs on the datasets.}
% 	\vspace{-4mm}
	\label{statistics}
\end{table}

%% file: sections/case_study.tex
\begin{table*}[]
	\centering
	\normalsize

	\resizebox{\textwidth}{!}{
	\begin{tabular}{{c}|{l}}
		\toprule[1.2pt]
		Code&\makecell[c]{Comment}\\
		\midrule[1.2pt]
		\multirow{7}{*}{\makecell{SQL: SELECT MAX(Capacity) FROM table\\WHERE Stadium = "Otkrytie Arena"}}
		&\textbf{Ground-Truth}: What is the maximum capacity of the Otkrytie Arena Stadium ?\\
		\cline{2-2}
		&Code-NN: What is the highest attendance for ?\\
		\cline{2-2}
		&P-G: Who is the \% that 's position at 51 ?\\
		\cline{2-2}
		&Tree2Seq: What is the highest capacity at $<$unk$>$ at arena ?\\
		\cline{2-2}
		&Graph2Seq: What is the highest capacity for arena arena ?\\
		\cline{2-2}
		&T2S+CP: What is the highest capacity for the stadium ?\\
		\cline{2-2}
		&TAG: What is the highest capacity for the stadium of Otkrytie Arena ?\\
		\midrule[1.2pt]
		\multirow{7}{*}{Python: {i: d [i] for i in d if i != 'c'}}
		&\textbf{Ground-Truth}: remove key 'c' from dictionary 'd'\\
		\cline{2-2}
		&Code-NN: remove all keys from a dictionary 'd'\\
		\cline{2-2}
		&P-G: select a string 'c' in have end of a list 'd'\\
		\cline{2-2}
		&Tree2Seq: get a key 'key' one ',' one ',' $<$unk$>$\\
		\cline{2-2}
		&\makecell[l]{Graph2Seq: filter a dictionary of dictionaries from a dictionary 'd'\\\qquad\qquad\quad where a dictionary of dictionaries 'd'}\\
		\cline{2-2}
		&T2S+CP: find all the values in dictionary 'd' from a dictionary 'd'\\
		\cline{2-2}
		&TAG: remove the key 'c' if a dictionary 'd'\\
		\bottomrule[1.2pt]
	\end{tabular}}
	\caption{Case study comparisons.}
	\label{case_study}
\end{table*}

%% file: sections/08_conclusion.tex
\section{Conclusion}
In this paper, we present a Type Auxiliary Guiding encoder-decoder framework for the code comment generation task. Our proposed framework takes full advantage of the type information associated with the code through the well designed \emph{Type-associated Encoder} and \emph{Type-restricted Decoder}. In addition, a hierarchical reinforcement learning method is provided for the training of our framework. 
The experimental results demonstrate significant improvements over state-of-the-art approaches and strong applicable potential in software development. 
Our proposed framework also verifies the necessity of the type information in the code translation related tasks with a practical framework and good results.
As future work, we will extend our framework to more complex contexts by devising efficient learning algorithms.

%% file: sections/acknowledgement.tex
\section*{Acknowledgments}
This research was supported in part by Natural Science Foundation of China (61876043, 61976052), Natural Science Foundation of Guangdong (2014A030306004, 2014A030308008), Science and Technology Planning Project of Guangzhou (201902010058). Besides, this project is also partly supported by the National Research Foundation, Prime Minister's Office, Singapore under its Campus for Research Excellence and Technological Enterprise (CREATE) programme.
This research was also made possible by NPRP grant NPRP10-0208-170408 from the Qatar National Research Fund (a member of Qatar Foundation). The findings herein reflect the work, and are solely the responsibility of the authors.

%% file: main.bbl
\begin{thebibliography}{38}
\expandafter\ifx\csname natexlab\endcsname\relax\def\natexlab#1{#1}\fi

\bibitem[{Aggarwal et~al.(2002)Aggarwal, Singh, and
  Chhabra}]{aggarwal2002integrated}
Krishan~K Aggarwal, Yogesh Singh, and Jitender~Kumar Chhabra. 2002.
\newblock An integrated measure of software maintainability.
\newblock In \emph{Annual Reliability and Maintainability Symposium. 2002
  Proceedings (Cat. No. 02CH37318)}, pages 235--241. IEEE.

\bibitem[{Allamanis et~al.(2016)Allamanis, Peng, and
  Sutton}]{allamanis2016convolutional}
Miltiadis Allamanis, Hao Peng, and Charles Sutton. 2016.
\newblock A convolutional attention network for extreme summarization of source
  code.
\newblock In \emph{International Conference on Machine Learning}, pages
  2091--2100.

\bibitem[{Alon et~al.(2018)Alon, Brody, Levy, and Yahav}]{alon2018code2seq}
Uri Alon, Shaked Brody, Omer Levy, and Eran Yahav. 2018.
\newblock code2seq: Generating sequences from structured representations of
  code.
\newblock \emph{arXiv preprint arXiv:1808.01400}.

\bibitem[{Cai et~al.(2018)Cai, Xu, Zhang, Yang, Li, and Liang}]{cai2018encoder}
Ruichu Cai, Boyan Xu, Zhenjie Zhang, Xiaoyan Yang, Zijian Li, and Zhihao Liang.
  2018.
\newblock An encoder-decoder framework translating natural language to database
  queries.
\newblock In \emph{Proceedings of the 27th International Joint Conference on
  Artificial Intelligence}, pages 3977--3983. AAAI Press.

\bibitem[{Dong and Lapata(2016)}]{dong-lapata-2016-language}
Li~Dong and Mirella Lapata. 2016.
\newblock \href {https://doi.org/10.18653/v1/P16-1004} {Language to logical
  form with neural attention}.
\newblock In \emph{Proceedings of the 54th Annual Meeting of the Association
  for Computational Linguistics (Volume 1: Long Papers)}, pages 33--43, Berlin,
  Germany. Association for Computational Linguistics.

\bibitem[{Eriguchi et~al.(2016)Eriguchi, Hashimoto, and
  Tsuruoka}]{eriguchi-etal-2016-tree}
Akiko Eriguchi, Kazuma Hashimoto, and Yoshimasa Tsuruoka. 2016.
\newblock \href {https://doi.org/10.18653/v1/P16-1078} {Tree-to-sequence
  attentional neural machine translation}.
\newblock In \emph{Proceedings of the 54th Annual Meeting of the Association
  for Computational Linguistics (Volume 1: Long Papers)}, pages 823--833,
  Berlin, Germany. Association for Computational Linguistics.

\bibitem[{Fernandes et~al.(2018)Fernandes, Allamanis, and
  Brockschmidt}]{fernandes2018structured}
Patrick Fernandes, Miltiadis Allamanis, and Marc Brockschmidt. 2018.
\newblock Structured neural summarization.
\newblock \emph{arXiv preprint arXiv:1811.01824}.

\bibitem[{Glorot and Bengio(2010)}]{glorot2010understanding}
Xavier Glorot and Yoshua Bengio. 2010.
\newblock Understanding the difficulty of training deep feedforward neural
  networks.
\newblock In \emph{Proceedings of the thirteenth international conference on
  artificial intelligence and statistics}, pages 249--256.

\bibitem[{Gu et~al.(2016)Gu, Lu, Li, and Li}]{gu-etal-2016-incorporating}
Jiatao Gu, Zhengdong Lu, Hang Li, and Victor~O.K. Li. 2016.
\newblock \href {https://doi.org/10.18653/v1/P16-1154} {Incorporating copying
  mechanism in sequence-to-sequence learning}.
\newblock In \emph{Proceedings of the 54th Annual Meeting of the Association
  for Computational Linguistics (Volume 1: Long Papers)}, pages 1631--1640,
  Berlin, Germany. Association for Computational Linguistics.

\bibitem[{Gumbel(1954)}]{gumbel1954statistical}
Emil~Julius Gumbel. 1954.
\newblock \emph{Statistical theory of extreme values and some practical
  applications: a series of lectures}, volume~33.
\newblock US Government Printing Office.

\bibitem[{Hu et~al.(2018)Hu, Li, Xia, Lo, and Jin}]{hu2018deep}
Xing Hu, Ge~Li, Xin Xia, David Lo, and Zhi Jin. 2018.
\newblock Deep code comment generation.
\newblock In \emph{Proceedings of the 26th Conference on Program
  Comprehension}, pages 200--210. ACM.

\bibitem[{Iyer et~al.(2016)Iyer, Konstas, Cheung, and
  Zettlemoyer}]{iyer-etal-2016-summarizing}
Srinivasan Iyer, Ioannis Konstas, Alvin Cheung, and Luke Zettlemoyer. 2016.
\newblock \href {https://doi.org/10.18653/v1/P16-1195} {Summarizing source code
  using a neural attention model}.
\newblock In \emph{Proceedings of the 54th Annual Meeting of the Association
  for Computational Linguistics (Volume 1: Long Papers)}, pages 2073--2083,
  Berlin, Germany. Association for Computational Linguistics.

\bibitem[{Jia and Liang(2016)}]{jia-liang-2016-data}
Robin Jia and Percy Liang. 2016.
\newblock \href {https://doi.org/10.18653/v1/P16-1002} {Data recombination for
  neural semantic parsing}.
\newblock In \emph{Proceedings of the 54th Annual Meeting of the Association
  for Computational Linguistics (Volume 1: Long Papers)}, pages 12--22, Berlin,
  Germany. Association for Computational Linguistics.

\bibitem[{Keneshloo et~al.(2019)Keneshloo, Shi, Ramakrishnan, and
  Reddy}]{keneshloo2019deep}
Yaser Keneshloo, Tian Shi, Naren Ramakrishnan, and Chandan~K Reddy. 2019.
\newblock Deep reinforcement learning for sequence-to-sequence models.
\newblock \emph{IEEE Transactions on Neural Networks and Learning Systems}.

\bibitem[{Kingma and Ba(2014)}]{kingma2014adam}
Diederik~P Kingma and Jimmy Ba. 2014.
\newblock Adam: A method for stochastic optimization.
\newblock \emph{arXiv preprint arXiv:1412.6980}.

\bibitem[{Li et~al.(2017)Li, Monroe, Shi, Jean, Ritter, and
  Jurafsky}]{li-etal-2017-adversarial}
Jiwei Li, Will Monroe, Tianlin Shi, S{\'e}bastien Jean, Alan Ritter, and Dan
  Jurafsky. 2017.
\newblock \href {https://doi.org/10.18653/v1/D17-1230} {Adversarial learning
  for neural dialogue generation}.
\newblock In \emph{Proceedings of the 2017 Conference on Empirical Methods in
  Natural Language Processing}, pages 2157--2169, Copenhagen, Denmark.
  Association for Computational Linguistics.

\bibitem[{Liang and Zhu(2018)}]{liang2018automatic}
Yuding Liang and Kenny~Qili Zhu. 2018.
\newblock Automatic generation of text descriptive comments for code blocks.
\newblock In \emph{Thirty-Second AAAI Conference on Artificial Intelligence}.

\bibitem[{Lin(2004)}]{lin2004rouge}
Chin-Yew Lin. 2004.
\newblock Rouge: A package for automatic evaluation of summaries.
\newblock In \emph{Text summarization branches out}, pages 74--81.

\bibitem[{Ling et~al.(2016)Ling, Blunsom, Grefenstette, Hermann,
  Ko{\v{c}}isk{\'y}, Wang, and Senior}]{ling-etal-2016-latent}
Wang Ling, Phil Blunsom, Edward Grefenstette, Karl~Moritz Hermann,
  Tom{\'a}{\v{s}} Ko{\v{c}}isk{\'y}, Fumin Wang, and Andrew Senior. 2016.
\newblock \href {https://doi.org/10.18653/v1/P16-1057} {Latent predictor
  networks for code generation}.
\newblock In \emph{Proceedings of the 54th Annual Meeting of the Association
  for Computational Linguistics (Volume 1: Long Papers)}, pages 599--609,
  Berlin, Germany. Association for Computational Linguistics.

\bibitem[{Movshovitz-Attias and
  Cohen(2013)}]{movshovitz-attias-cohen-2013-natural}
Dana Movshovitz-Attias and William~W. Cohen. 2013.
\newblock \href {https://www.aclweb.org/anthology/P13-2007} {Natural language
  models for predicting programming comments}.
\newblock In \emph{Proceedings of the 51st Annual Meeting of the Association
  for Computational Linguistics (Volume 2: Short Papers)}, pages 35--40, Sofia,
  Bulgaria. Association for Computational Linguistics.

\bibitem[{Ng et~al.(1999)Ng, Harada, and Russell}]{ng1999policy}
Andrew~Y Ng, Daishi Harada, and Stuart Russell. 1999.
\newblock Policy invariance under reward transformations: Theory and
  application to reward shaping.
\newblock In \emph{ICML}, volume~99, pages 278--287.

\bibitem[{Papineni et~al.(2002)Papineni, Roukos, Ward, and
  Zhu}]{papineni-etal-2002-bleu}
Kishore Papineni, Salim Roukos, Todd Ward, and Wei-Jing Zhu. 2002.
\newblock \href {https://doi.org/10.3115/1073083.1073135} {{B}leu: a method for
  automatic evaluation of machine translation}.
\newblock In \emph{Proceedings of the 40th Annual Meeting of the Association
  for Computational Linguistics}, pages 311--318, Philadelphia, Pennsylvania,
  USA. Association for Computational Linguistics.

\bibitem[{Rabinovich et~al.(2017)Rabinovich, Stern, and
  Klein}]{rabinovich-etal-2017-abstract}
Maxim Rabinovich, Mitchell Stern, and Dan Klein. 2017.
\newblock \href {https://doi.org/10.18653/v1/P17-1105} {Abstract syntax
  networks for code generation and semantic parsing}.
\newblock In \emph{Proceedings of the 55th Annual Meeting of the Association
  for Computational Linguistics (Volume 1: Long Papers)}, pages 1139--1149,
  Vancouver, Canada. Association for Computational Linguistics.

\bibitem[{Ranzato et~al.(2015)Ranzato, Chopra, Auli, and
  Zaremba}]{ranzato2015sequence}
Marc'Aurelio Ranzato, Sumit Chopra, Michael Auli, and Wojciech Zaremba. 2015.
\newblock Sequence level training with recurrent neural networks.
\newblock \emph{arXiv preprint arXiv:1511.06732}.

\bibitem[{See et~al.(2017)See, Liu, and Manning}]{see-etal-2017-get}
Abigail See, Peter~J. Liu, and Christopher~D. Manning. 2017.
\newblock \href {https://doi.org/10.18653/v1/P17-1099} {Get to the point:
  Summarization with pointer-generator networks}.
\newblock In \emph{Proceedings of the 55th Annual Meeting of the Association
  for Computational Linguistics (Volume 1: Long Papers)}, pages 1073--1083,
  Vancouver, Canada. Association for Computational Linguistics.

\bibitem[{Tai et~al.(2015)Tai, Socher, and Manning}]{tai-etal-2015-improved}
Kai~Sheng Tai, Richard Socher, and Christopher~D. Manning. 2015.
\newblock \href {https://doi.org/10.3115/v1/P15-1150} {Improved semantic
  representations from tree-structured long short-term memory networks}.
\newblock In \emph{Proceedings of the 53rd Annual Meeting of the Association
  for Computational Linguistics and the 7th International Joint Conference on
  Natural Language Processing (Volume 1: Long Papers)}, pages 1556--1566,
  Beijing, China. Association for Computational Linguistics.

\bibitem[{Tenny(1988)}]{tenny1988program}
Ted Tenny. 1988.
\newblock Program readability: Procedures versus comments.
\newblock \emph{IEEE Transactions on Software Engineering}, 14(9):1271--1279.

\bibitem[{Vinyals et~al.(2015)Vinyals, Fortunato, and
  Jaitly}]{vinyals2015pointer}
Oriol Vinyals, Meire Fortunato, and Navdeep Jaitly. 2015.
\newblock Pointer networks.
\newblock In \emph{Advances in Neural Information Processing Systems}, pages
  2692--2700.

\bibitem[{Wan et~al.(2018)Wan, Zhao, Yang, Xu, Ying, Wu, and
  Yu}]{wan2018improving}
Yao Wan, Zhou Zhao, Min Yang, Guandong Xu, Haochao Ying, Jian Wu, and Philip~S
  Yu. 2018.
\newblock Improving automatic source code summarization via deep reinforcement
  learning.
\newblock In \emph{Proceedings of the 33rd ACM/IEEE International Conference on
  Automated Software Engineering}, pages 397--407. ACM.

\bibitem[{Wong and Mooney(2007)}]{wong-mooney-2007-generation}
Yuk~Wah Wong and Raymond Mooney. 2007.
\newblock \href {https://www.aclweb.org/anthology/N07-1022} {Generation by
  inverting a semantic parser that uses statistical machine translation}.
\newblock In \emph{Human Language Technologies 2007: The Conference of the
  North {A}merican Chapter of the Association for Computational Linguistics;
  Proceedings of the Main Conference}, pages 172--179, Rochester, New York.
  Association for Computational Linguistics.

\bibitem[{Wu et~al.(2018{\natexlab{a}})Wu, Tian, Qin, Lai, and
  Liu}]{wu-etal-2018-study}
Lijun Wu, Fei Tian, Tao Qin, Jianhuang Lai, and Tie-Yan Liu.
  2018{\natexlab{a}}.
\newblock \href {https://doi.org/10.18653/v1/D18-1397} {A study of
  reinforcement learning for neural machine translation}.
\newblock In \emph{Proceedings of the 2018 Conference on Empirical Methods in
  Natural Language Processing}, pages 3612--3621, Brussels, Belgium.
  Association for Computational Linguistics.

\bibitem[{Wu et~al.(2018{\natexlab{b}})Wu, Xia, Tian, Zhao, Qin, Lai, and
  Liu}]{wu2018adversarial}
Lijun Wu, Yingce Xia, Fei Tian, Li~Zhao, Tao Qin, Jianhuang Lai, and Tie-Yan
  Liu. 2018{\natexlab{b}}.
\newblock Adversarial neural machine translation.
\newblock In \emph{Asian Conference on Machine Learning}, pages 534--549.

\bibitem[{Wu et~al.(2016)Wu, Schuster, Chen, Le, Norouzi, Macherey, Krikun,
  Cao, Gao, Macherey et~al.}]{wu2016google}
Yonghui Wu, Mike Schuster, Zhifeng Chen, Quoc~V Le, Mohammad Norouzi, Wolfgang
  Macherey, Maxim Krikun, Yuan Cao, Qin Gao, Klaus Macherey, et~al. 2016.
\newblock Google's neural machine translation system: Bridging the gap between
  human and machine translation.
\newblock \emph{arXiv preprint arXiv:1609.08144}.

\bibitem[{Xu et~al.(2018{\natexlab{a}})Xu, Wu, Wang, Feng, and
  Sheinin}]{xu-etal-2018-sql}
Kun Xu, Lingfei Wu, Zhiguo Wang, Yansong Feng, and Vadim Sheinin.
  2018{\natexlab{a}}.
\newblock \href {https://doi.org/10.18653/v1/D18-1112} {{SQL}-to-text
  generation with graph-to-sequence model}.
\newblock In \emph{Proceedings of the 2018 Conference on Empirical Methods in
  Natural Language Processing}, pages 931--936, Brussels, Belgium. Association
  for Computational Linguistics.

\bibitem[{Xu et~al.(2018{\natexlab{b}})Xu, Wu, Wang, Feng, Witbrock, and
  Sheinin}]{xu2018graph2seq}
Kun Xu, Lingfei Wu, Zhiguo Wang, Yansong Feng, Michael Witbrock, and Vadim
  Sheinin. 2018{\natexlab{b}}.
\newblock Graph2seq: Graph to sequence learning with attention-based neural
  networks.
\newblock \emph{arXiv preprint arXiv:1804.00823}.

\bibitem[{Yin et~al.(2018)Yin, Deng, Chen, Vasilescu, and
  Neubig}]{yin2018learning}
Pengcheng Yin, Bowen Deng, Edgar Chen, Bogdan Vasilescu, and Graham Neubig.
  2018.
\newblock Learning to mine aligned code and natural language pairs from stack
  overflow.
\newblock In \emph{2018 IEEE/ACM 15th International Conference on Mining
  Software Repositories (MSR)}, pages 476--486. IEEE.

\bibitem[{Yin and Neubig(2017)}]{yin-neubig-2017-syntactic}
Pengcheng Yin and Graham Neubig. 2017.
\newblock \href {https://doi.org/10.18653/v1/P17-1041} {A syntactic neural
  model for general-purpose code generation}.
\newblock In \emph{Proceedings of the 55th Annual Meeting of the Association
  for Computational Linguistics (Volume 1: Long Papers)}, pages 440--450,
  Vancouver, Canada. Association for Computational Linguistics.

\bibitem[{Zhong et~al.(2017)Zhong, Xiong, and Socher}]{zhong2017seq2sql}
Victor Zhong, Caiming Xiong, and Richard Socher. 2017.
\newblock Seq2sql: Generating structured queries from natural language using
  reinforcement learning.
\newblock \emph{arXiv preprint arXiv:1709.00103}.

\end{thebibliography}
